\documentclass[11pt]{article}

\usepackage[final]{acl}

\usepackage{times}
\usepackage{latexsym}
\usepackage{algorithm}
\usepackage{algorithmic}
\usepackage{amsmath}
\usepackage{amssymb}
\usepackage{subcaption}

\usepackage{booktabs}
\usepackage{multirow}
\usepackage{array}

\usepackage{cuted}   
\usepackage{caption} 
\usepackage{listings}
\usepackage{listingsutf8}
\usepackage[most]{tcolorbox}
\tcbuselibrary{breakable,listings}

\lstdefinestyle{promptstyle}{
  basicstyle=\ttfamily\footnotesize,
  columns=fullflexible,
  keepspaces=true,
  breaklines=true,
  showstringspaces=false,
    extendedchars=\true,
    inputencoding=utf8
}

\newtcblisting{promptbox}[2][]{%
  breakable,
  listing only,
  title={#2},
  fonttitle=\bfseries,
  colback=white,
  colframe=black,
  left=1mm,right=1mm,top=1mm,bottom=1mm,
  listing options={style=promptstyle},
  #1
}

\usepackage[T1]{fontenc}

\usepackage[utf8]{inputenc}

\usepackage{microtype}

\usepackage{inconsolata}

\usepackage{graphicx}

\title{Controlled Self-Evolution for Algorithmic Code Optimization}

\author{
\textbf{Tu Hu\textsuperscript{1*}},
\textbf{Ronghao Chen\textsuperscript{2,7*}},
\textbf{Shuo Zhang\textsuperscript{3*}},
\textbf{Jianghao Yin\textsuperscript{4}},
\textbf{Mou Xiao Feng\textsuperscript{3}},
\textbf{Jingping Liu\textsuperscript{5}},
\\
\textbf{Shaolei Zhang\textsuperscript{6}},
\textbf{Andy Wang\textsuperscript{8}},
\textbf{Wenqi Jiang\textsuperscript{1}},
\textbf{Yuqi Fang\textsuperscript{1\dag}},
\textbf{Sen Hu\textsuperscript{2,7\dag}},
\textbf{Huacan Wang\textsuperscript{3\dag}},
\textbf{Yi Xu\textsuperscript{3\dag}}
\\
\\
 \textsuperscript{1}NJU,
 \textsuperscript{2}PKU,
 \textsuperscript{3}Midea-AIRC,
 \textsuperscript{4}ECNU,
 \textsuperscript{5}SYSU,
 \textsuperscript{6}RUC,
 \textsuperscript{7}QuantaAlpha,
 \textsuperscript{8}QuantML
\\
\small {
    \textbf{\textsuperscript{*}These authors contributed equally to this work.}
}
\\
 \small{
   \textbf{\dag Correspondence:} 
   \href{yqfang@nju.edu.cn}{yqfang@nju.edu.cn},
   \href{mailto:husen@pku.edu.cn}{husen@pku.edu.cn},
   \href{mailto:wanghc141@midea.com}{wanghc141@midea.com}, 
   \href{mailto:xuyi42@midea.com}{xuyi42@midea.com}
 }
}

\begin{document}
\maketitle

\begin{abstract}
Self-evolution methods enhance code generation through iterative ``generate-verify-refine'' cycles, yet existing approaches suffer from low exploration efficiency, failing to discover solutions with superior complexity within limited budgets. This inefficiency stems from initialization bias trapping evolution in poor solution regions, uncontrolled stochastic operations lacking feedback guidance, and insufficient experience utilization across tasks.
To address these bottlenecks, we propose \textbf{C}ontrolled \textbf{S}elf-\textbf{E}volution (CSE), which consists of three key components. Diversified Planning Initialization generates structurally distinct algorithmic strategies for broad solution space coverage. Genetic Evolution replaces stochastic operations with feedback-guided mechanisms, enabling targeted mutation and compositional crossover. Hierarchical Evolution Memory captures both successful and failed experiences at inter-task and intra-task levels.
Experiments on EffiBench-X demonstrate that CSE consistently outperforms all baselines across various LLM backbones. Furthermore, CSE achieves higher efficiency from early generations and maintains continuous improvement throughout evolution.
Our code is publicly available at \url{https://github.com/QuantaAlpha/EvoControl}.
\end{abstract}

\begin{figure}[h]
  \centering

  \begin{subfigure}[t]{0.7\linewidth}
    \centering
    \includegraphics[width=\linewidth]{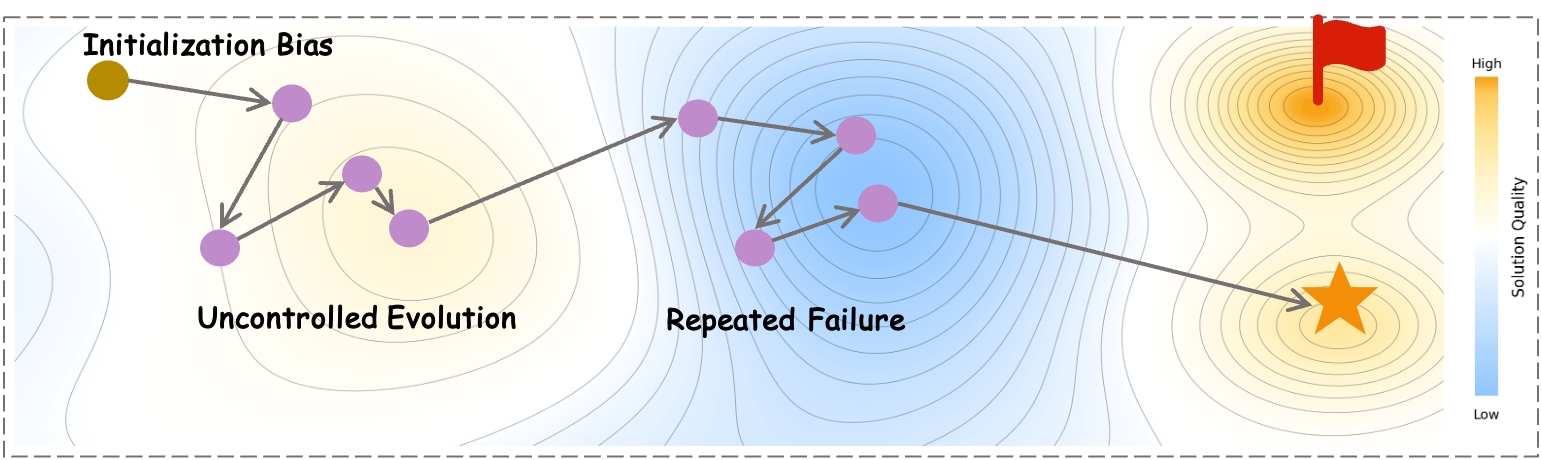}
    \caption{Existing Method.}
    \label{fig:sub1}
  \end{subfigure}

  \begin{subfigure}[t]{0.7\linewidth}
    \centering
    \includegraphics[width=\linewidth]{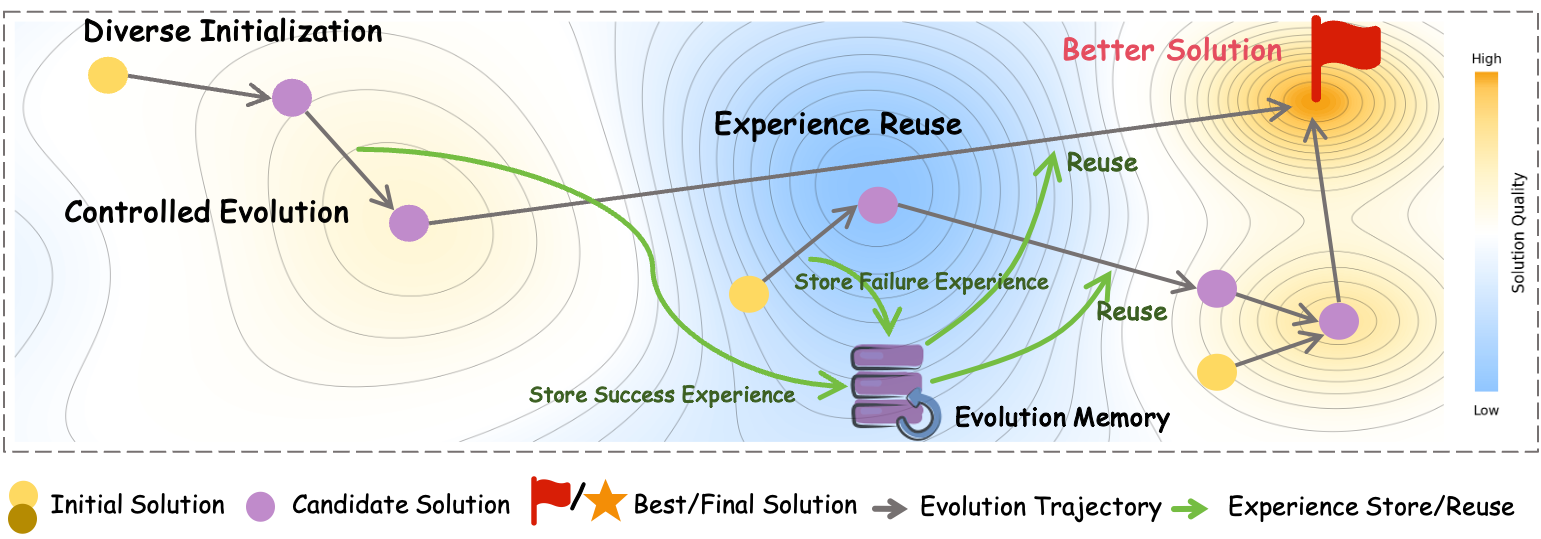}
    \caption{Controlled Self-Evolution (CSE).}
    \label{fig:sub2}
  \end{subfigure}

    \caption{\textbf{Controlled Self-Evolution improves exploration efficiency.}
    (a) Existing self-evolution wastes budget on low-quality regions due to initialization bias, uncontrolled evolution and repeated failure.
    (b) CSE guides exploration toward higher-quality solutions through diversified initialization, controlled evolution, and experience reuse.}
    \label{fig:intro}
\end{figure}

\section{Introduction}

Code generation has emerged as a critical application of Large Language Models (LLMs) \cite{gemini2.5, gpt4,internlm2,llam3, kimi15,deepseekr1}, with models demonstrating impressive capabilities in producing functionally correct solutions for programming tasks \cite{humaneval,codeLLMSurvey, CodeAgentsurvey, codeagentsurvey2, qwen2.5coder, code1, nips_code1}. Early approaches relied on single-turn generation, where models directly produce complete solutions from problem specifications \cite{wizardcoder, magicoder}. While achieving reasonable success on simple tasks, this paradigm struggles with complex algorithmic problems and lacks mechanisms to interact with execution environments or leverage verification feedback. To address these limitations, self-evolution methods \cite{self-evolvingsurvey, self-evolvingsurvey2} have emerged as a promising paradigm that enables iterative ``generate-verify-refine'' cycles: models start from initial solutions, execute code against test cases, analyze feedback signals (e.g., failed tests, performance bottlenecks), and generate improved variants. Methods such as AlphaEvolve \cite{novikov2025alphaevolve} and SE-Agent \cite{Se-agent} have demonstrated that treating code generation as a feedback-driven search process can substantially enhance code quality.

Ideally, with unlimited computational resources, we could allow these self-evolving agents to explore extensively until reaching optimal solutions that are not only functionally correct but also exhibit superior time and space complexity. However, practical deployment scenarios impose strict resource constraints. In real-world applications, extensive multi-turn inference incurs prohibitive computational costs and latency. This creates a fundamental tension: we need methods that can discover high-quality solutions, measured not just by correctness but by algorithmic efficiency, within a limited exploration budget.

Unfortunately, existing self-evolution methods suffer from critically low exploration efficiency. This inefficiency prevents them from discovering code with superior time and space complexity within limited exploration budgets, as the search process fails to efficiently navigate toward algorithmically optimal solutions. This inefficiency stems from three fundamental limitations (as shown in Figure \ref{fig:intro}). First, \textit{initialization bias}: methods \cite{afterburner,effilearner} typically begin evolution from a single or few initial solutions generated by the base model, which may lie in poor regions of the solution space, necessitating many iterations to escape local optima. Second, \textit{uncontrolled stochastic evolution}: operations like random mutation and crossover are applied without explicit guidance from feedback signals, resulting in undirected exploration where many generated variants fail to systematically navigate toward better solutions. Third, \textit{insufficient utilization of evolution experience}: existing approaches \cite{Se-agent, afterburner} do not effectively accumulate successful patterns within a task or abstract transferable experiences across tasks, causing repeated failures and preventing the reuse of proven optimization strategies.

This motivates our central research question: \textbf{Can we design a self-evolution framework that achieves high code quality while dramatically improving exploration efficiency?} We propose \textbf{Controlled Self-Evolution (CSE)}, a novel framework that addresses all three efficiency bottlenecks through three key innovations. First, Diversified Planning Initialization generates multiple structurally distinct algorithmic strategies before evolution begins, ensuring broad coverage of the solution space and reducing the likelihood of getting trapped in poor local regions. Second, Genetic Evolution replaces stochastic operations with fine-grained feedback-guided mechanisms: functional decomposition enables targeted mutation that refines faulty components while preserving high-performing parts, and compositional crossover structurally integrates complementary strengths from diverse solutions. Third, Hierarchical Evolution Memory captures and reuses evolutionary insights at two levels: local memory accumulates task-specific lessons to avoid repeating failures within the current problem, while global memory distills cross-task optimization patterns into reusable templates that accelerate future evolution. 
We conduct comprehensive experiments on EffiBench-X\cite{effibench}, showing that CSE consistently outperforms all baselines across diverse LLM backbones, achieves stronger efficiency from early generations, and continues improving throughout evolution. These results demonstrate robust, backbone-agnostic gains with both fast-starting and sustained progress.
\section{Related Work}

\paragraph{Code Generation with LLMs.}
Large Language Models have demonstrated remarkable capabilities in code generation \cite{CodeAgentsurvey,nips_code2,nips_code3,nips_code4,nips_code5,nips_code6}. Recent advances span instruction tuning (WizardCoder \cite{wizardcoder}, Magicoder \cite{magicoder}), retrieval augmentation \cite{reacc}, while specialized models such as GPT-4o \cite{gpt4o} and DeepSeek-Coder \cite{deepseekcoder} have achieved strong results on programming benchmarks. However, recent evaluations consistently reveal that LLMs tend to generate solutions that are "correct yet inefficient." For instance, EffiBench-X \cite{effibench} highlights a significant efficiency gap between model-generated code and human-written canonical solutions. These findings underscore that code generation capability does not equate to evolution proficiency, necessitating the introduction of post-generation systematic refinement mechanisms.

\paragraph{Self-Evolution.}
To address single-turn generation limitations, self-evolution methods enable iterative refinement through "generate-verify-refine" cycles. Self-reflection \cite{Self-refine,reflexion,afterburner,adaplanner} approaches enable models to learn from execution feedback, though they focus on debugging rather than algorithmic optimization. Such as AfterBurner \cite{afterburner} which perform local refinement but suffer from initialization bias and local optima.  Population-based approaches \cite{FunSearch, EOH, shinkaevolve} such as AlphaEvolve \cite{novikov2025alphaevolve} employ evolutionary resampling with stochastic mutations, while SE-Agent \cite{Se-agent} introduces trajectory-level evolution via step-wise recombination. However, these methods face low search efficiency due to unguided exploration through random mutations, and lack of experience reuse. They thus require extensive iterations and often fail to discover optimal solutions within limited budgets. Our CSE framework addresses this through diversified initialization, feedback-guided operations, and hierarchical memory for efficient, controlled evolution.
\section{Problem Formulation}

We formalize the algorithmic code efficiency optimization task as follows. Given a problem specification $x$ describing functional requirements and constraints, the goal is to generate an implementation $y$ that satisfies correctness while achieving optimal execution efficiency.

The optimization environment is defined as $\mathcal{O} = (\mathcal{X}, \mathcal{Y}, \mathcal{F})$, where $\mathcal{X}$ represents the problem specification space, $\mathcal{Y}$ denotes the solution space, and $\mathcal{F}: \mathcal{Y} \times \mathcal{X} \rightarrow \mathbb{R}$ is a reward function evaluating solution quality. The reward function captures both correctness and efficiency.

The optimization unfolds as a population-based evolutionary process over $T$ iterations. At each step $t$, we maintain a population $\mathcal{P}_t = \{y^{(t)}_1, y^{(t)}_2, \ldots, y^{(t)}_{N_t}\}$ of $N_t$ candidate solutions. The evolutionary trajectory $\mathcal{T} = (\mathcal{P}_0, \mathcal{P}_1, \ldots, \mathcal{P}_T)$ progresses through feedback-driven selection, mutation, and crossover operators. Our objective is to discover the optimal solution $y^* = \arg\max_{y \in \bigcup_{t=0}^{T} \mathcal{P}_t} \mathcal{F}(y, x)$, requiring efficient navigation of the solution space through controlled exploration rather than uncontrolled stochastic evolution.
\section{Method}

\begin{figure*}[thb!]
  \centering
  \includegraphics[width=1\textwidth]{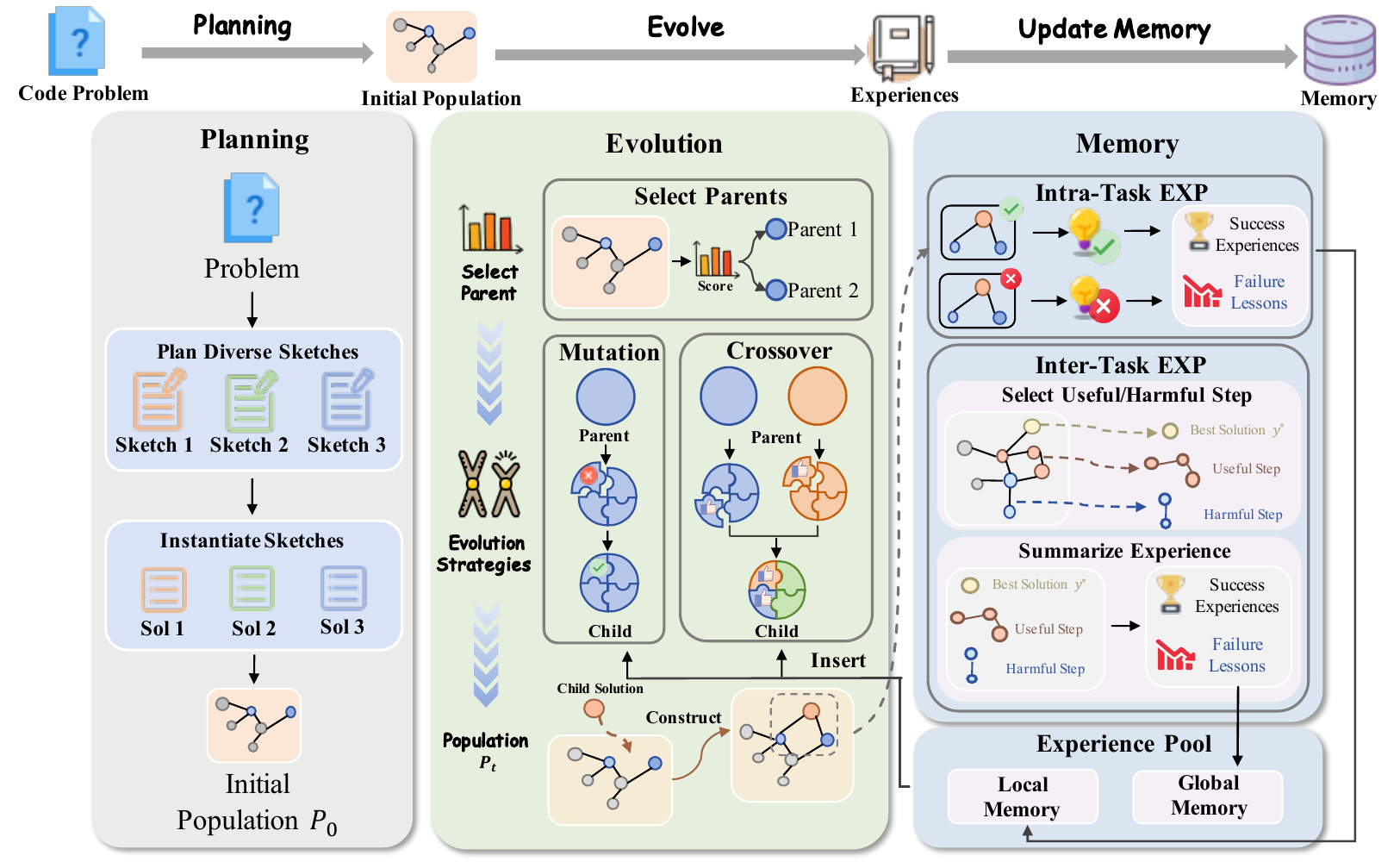}\\
  \caption{\textbf{Overview of the CSE.} Our method consists of three key components: Diversified Planning Initialization, Genetic Evolution, and Hierarchical Evolution Memory.}
  \label{fig:framework}
\end{figure*}

As illustrated in Figure \ref{fig:framework}, our method consists of three key components: \textbf{Diversified Planning Initialization}, \textbf{Genetic Evolution}, and \textbf{Hierarchical Evolution Memory}. 
Diversified Planning Initialization generates high-quality and diverse initial solutions through varied planning strategies, providing a strong foundation for subsequent evolution. Unlike prior black-box evolutionary approaches, Genetic Evolution enables fine-grained, controllable mutation to efficiently navigate toward optimal solutions. Hierarchical Evolution Memory hierarchically summarizes both intra-task and inter-task experiences, providing search guidance for the evolutionary process.
In the following subsections, we describe each component in detail.

\subsection{Diversified Planning Initialization}

To maximize the diversity of initial solutions and ensure broad coverage of the solution space, we employ a two-stage approach: diverse planning followed by completion. Specifically, we prompt the agent $\mathcal{A}_{\theta}$ with explicit diversity constraints to generate a set of high-level solution sketches $\mathcal{Z} = \{z_1, \ldots, z_{N_\text{init}}\}$, where each sketch represents a semantically distinct strategy. For example, given a code generation task, the agent might propose fundamentally different approaches such as a greedy algorithm, dynamic programming, or bit manipulation optimization, rather than superficial variations of the same logic.

Subsequently, each solution sketch is instantiated into a concrete code implementation, forming the initial candidate solutions:
\begin{equation}
y_i^{(0)} \sim \mathcal{A}_{\theta}(y \mid x, z_i), \quad \forall i \in \{1,\ldots,N_\text{init}\},
\end{equation}
where $y_i^{(0)}$ denotes the complete output generated by the model conditioned on the input specification $x$ and solution sketch $z_i$ at the initial stage. Through this approach, the initial population $\mathcal{P}_0=\{y_1^{(0)}, . . . , y_{N_\text{init}}^{(0)} \}$ is no longer composed of random perturbations around a single mode, but instead achieves diverse coverage of the solution space. This ensures that the subsequent evolutionary process can explore multiple promising regions in parallel, significantly reducing the risk of premature convergence to a suboptimal local optimum.

\subsection{Genetic Evolution}

The goal of self-evolutionary update is to guide the model in searching for better trajectories based on existing ones. A key insight is that incorporating explicit control into the search process, rather than relying on uncontrolled stochastic evolution, can significantly improve search efficiency. To this end, we design a Genetic Evolution mechanism with the following components:

\paragraph{Parent Selection.} We depart from prior evolutionary approaches \cite{Se-agent} that exclusively select high-reward solutions, as low-reward solutions may also contain valuable parts. We design a probability-based parent selection strategy where the selection probability of each candidate is proportional to its normalized reward:
\begin{equation}
P_{\text{select}}(y_i^{(t)}) = \frac{\mathcal{F}(y_i^{(t)}, x)}{\sum_{j=1}^{|\mathcal{P}_t|} \mathcal{F}(y_j^{(t)}, x)},
\label{eq:select}
\end{equation}
where $P_{\text{select}}(y_i^{(t)})$ denotes the selection probability of candidate solution $y_i^{(t)}$, and $\mathcal{P}_t$ represents the population at evolution step $t$. This mechanism implements a soft selection distribution that prioritizes high-reward individuals while retaining lower-reward individuals that may contain useful logic fragments or partial solutions for recombination.

\paragraph{Evolution Strategies.} To enable fine-grained controlled evolution, we first prompt the agent to decompose the code $y$ into a set of disjoint functional components $\{c_1, c_2, \ldots, c_m\}$ (e.g., I/O parsing module, core algorithm logic, boundary condition handling). This functional decomposition provides the necessary structural foundation for subsequent fine-grained interventions. We then equip the agent with two controlled evolution strategies:

\textbf{Controlled Mutation.} The agent employs self-reflection to identify the specific faulty component $c_{\text{faulty}}$ responsible for low reward. After localizing the problematic module, we perform targeted regeneration on $c_{\text{faulty}}$ while keeping the remaining $m-1$ well-performing components frozen:
\begin{equation}
\begin{split}
    y_{\text{child}} = \{c_1, \ldots, \text{Refine}(c_{\text{faulty}}),\ldots, c_m\}.
\end{split}
\label{eq:refine}
\end{equation}
This surgical repair strategy not only avoids disruptive interference with the global context but also significantly improves mutation efficiency.

\textbf{Compositional Crossover.} To facilitate the flow of advantageous traits across the population, we introduce compositional crossover, which performs logic-level recombination of strengths rather than naive textual concatenation. Suppose parent solutions $y_A$ and $y_B$ exhibit complementary advantages in different components (e.g., $y_A$ has superior time complexity while $y_B$ demonstrates better robustness). The crossover operator synthesizes these strengths structurally:
\begin{equation}
y_{\text{child}} = \text{Crossover}(\{c_{\text{time}}^{(A)}\}, \{c_{\text{robust}}^{(B)}\}),
\label{eq:crossover}
\end{equation}
where $c_{\text{time}}^{(A)}$ denotes the time-efficient component from solution $y_A$, and $c_{\text{robust}}^{(B)}$ represents the robustness-oriented component from solution $y_B$. This enables the agent to mimic the human practice of combining complementary strengths from different solutions.

\subsection{Hierarchical Evolution Memory}

To effectively leverage both intra-task and inter-task experiences, we propose Hierarchical Evolution Memory. 

\paragraph{Local Memory.}
$\mathcal{M}_{\text{local}}$ aims to capture immediate experiences from intra-task search trajectories. At each evolutionary step $t$, we compare the reward change $\Delta_t \;=\; \mathcal{F}(y^{(t)}_{\mathrm{child}},x) - \mathcal{F}(y^{(t)}_{\mathrm{parent}},x)$ between the parent $y^{(t)}_{\mathrm{parent}}$ and its child $y^{(t)}_{\mathrm{child}}$.
The agent extracts two types of critical experiences: success insights ($\Delta_t > 0$) that analyze what led to improvements and are marked as positive patterns to retain, and failure lessons ($\Delta_t \le 0$) that analyze causes of reward degradation and are marked as negative constraints to avoid. This reflection process is formalized as:
\begin{subequations}
\label{eq:local}
\begin{align}
m_t &\leftarrow \text{Reflect}\!\left(y^{(t)}_{\mathrm{child}},\,y^{(t)}_{\mathrm{parent}},\,\Delta_t\right),
\label{eq:local_reflect}\\
\mathcal{M}^{\text{loc}}_t &\leftarrow \text{Compress}\!\left(\mathcal{M}^{\text{loc}}_{t-1}\cup\{m_t\}\right),
\label{eq:local_compress}
\end{align}
\end{subequations}
where $m_t$ denotes the distilled experience at step $t$. At evolution step $t+1$, the accumulated $\mathcal{M}_{\text{local}}$ is dynamically injected into the prompt context, forming bidirectional guidance: success insights explicitly instruct the model to reuse validated optimization strategies, while failure lessons prevent the model from repeating known mistakes. To prevent memory overflow as iterations accumulate, whenever $\mathcal{M}^{\text{loc}}_t$ exceeds a predefined length threshold, we perform semantic compression to maintain high information density in the local memory.

\paragraph{Global Memory.}
$\mathcal{M}_{\text{glb}}$ aims to distill and reuse inter-task experiences. For each task $\tau$, we collect all evolution steps and their reward changes.
We keep the top-$K$ improving useful steps and top-$K$ degrading harmful steps (ranked by $\Delta$),
denoted by $\mathcal{S}^{+}_\tau$ and $\mathcal{S}^{-}_\tau$, respectively.
The LLM distills a task-level global experience $g_\tau$ with $\mathcal{S}^{+}_\tau$, $\mathcal{S}^{-}_\tau$ and best solution $y^*_\tau$ and stores it in a vector database:
\begin{subequations}
\label{eq:global}
\begin{align}
g_\tau &\leftarrow \text{Reflect}\!\left(\mathcal{S}^{+}_\tau,\,\mathcal{S}^{-}_\tau,\,y^*_\tau\right),
\label{eq:global_reflect}\\
\mathcal{M}^{\text{glb}} &\leftarrow \mathcal{M}^{\text{glb}} \cup \{g_\tau\}.
\label{eq:global_update}
\end{align}
\end{subequations}
At evolutionary step $t$, the agent  generates $N_{\text{q}}$ targeted retrieval queries based on the current evolutionary context (including the current code state, encountered errors, or performance bottlenecks):
\begin{subequations}
\label{eq:retrieve}
\begin{align}
\{q_{t}^{(n)}\}_{n=1}^{N_q} &\leftarrow \text{GenerateQueries}(\text{Context}_t),
\label{eq:retrieve_queries}\\
\mathcal{E}_t^{\text{ret}} &\leftarrow \text{Retrieve}(\mathcal{M}^{\text{glb}}, \{q_t^{(n)}\}_{n=1}^{N_q}),
\label{eq:retrieve_exec}
\end{align}
\end{subequations}
where $\{q_{t}^{(n)}\}_{n=1}^{N_q}$ represents the generated queries and $\mathcal{E}_t^{\text{ret}}$ denotes the retrieved relevant experiences. This mechanism ensures that the agent can precisely invoke past experiences from similar tasks when encountering specific challenges.

As summarized in Algorithm~\ref{alg:cse}, CSE first generates diverse strategy sketches $\mathcal{Z}$ and instantiates them into an initial population $\mathcal{P}_0$ with $\mathcal{A}_\theta$. 
It then iterates for $T$ steps: at iteration $t$, it samples parent solution(s) from $\mathcal{P}_{t-1}$ via probabilistic selection (Eq.~\ref{eq:select}), retrieves relevant experiences $\mathcal{E}^{\text{ret}}_t$ from global memory $\mathcal{M}^{\text{glb}}$, and composes them with local memory $\mathcal{M}^{\text{loc}}_{t-1}$ as context. 
Conditioned on this context, the agent applies controlled mutation (Eq.~\ref{eq:refine}) or compositional crossover (Eq.~\ref{eq:crossover}) to generate an offspring, evaluates it by $\mathcal{F}(\cdot,x)$, and updates $\mathcal{P}_t$ and $\mathcal{M}^{\text{loc}}_t$ using the reward change $\Delta_t$ (Eq.~\ref{eq:local}). 
Finally, CSE returns the best solution $y^*$ and updates $\mathcal{M}^{\text{glb}}$ by distilling intra-task experiences $g_\tau$ from top-$K$ improving and degrading steps (Eq.~\ref{eq:global}).

\section{Experiments}

\begin{table*}[!t]
\centering
\small
\setlength{\tabcolsep}{6pt}

\begin{tabular}{lccccccccc}
\toprule
& \multicolumn{3}{c}{\textbf{Python}} & \multicolumn{3}{c}{\textbf{C++}} & \multicolumn{3}{c}{\textbf{Avg}} \\
\cmidrule(lr){2-4} \cmidrule(lr){5-7} \cmidrule(lr){8-10}
& ET & MP & MI & ET & MP & MI & ET & MP & MI \\
\midrule

\multicolumn{10}{c}{\textbf{Qwen3-235B-A22B}} \\
\midrule
Direct          & 31.10\% & 49.13\% & 46.28\% & 17.42\% & 46.55\% & 19.24\% & 24.26\% & 47.84\% & 32.76\% \\
Self-Reflection & 48.49\% & 49.82\% & 50.89\% & 47.47\% & 48.00\% & 44.75\% & 47.98\% & 48.91\% & 47.82\% \\
SE-Agent        & 48.57\% & 50.39\% & 49.63\% & \textbf{50.71\%} & 48.25\% & 47.19\% & \textbf{49.64\%} & 49.32\% & 48.41\% \\
AlphaEvolve     & 48.70\% & 50.33\% & 51.47\% & 48.75\% & 48.67\% & 46.58\% & 48.73\% & 49.50\% & 49.03\% \\
CSE (Ours)      & \textbf{49.06\%} & \textbf{50.99\%} & \textbf{52.38\%}
                & 48.70\% & \textbf{49.03\%} & \textbf{47.95\%} & 48.88\% & \textbf{50.01\%} & \textbf{50.17\%} \\
\midrule

\multicolumn{10}{c}{\textbf{DeepSeek-v3-0324}} \\
\midrule
Direct          & 52.90\% & 58.89\% & 56.22\% & 44.21\% & 45.79\% & 41.01\% & 48.56\% & 52.34\% & 48.62\% \\
Self-Reflection & 57.35\% & 60.20\% & 61.18\% & 45.43\% & 46.53\% & 43.16\% & 51.39\% & 53.37\% & 52.17\% \\
SE-Agent        & 60.09\% & 59.68\% & 60.69\% & 47.74\% & 46.42\% & 44.54\% & 53.92\% & 53.05\% & 52.62\% \\
AlphaEvolve     & 58.86\% & 60.04\% & 60.24\% & 46.19\% & 46.41\% & 44.02\% & 52.53\% & 53.23\% & 52.13\% \\
CSE (Ours)      & \textbf{61.06\%} & \textbf{61.03\%} & \textbf{62.80\%}
                & \textbf{48.75\%} & \textbf{46.89\%} & \textbf{47.37\%} & \textbf{54.91\%} & \textbf{53.96\%} & \textbf{55.09\%} \\
\midrule

\multicolumn{10}{c}{\textbf{Claude-4.5-Sonnet}} \\
\midrule
Direct          & 65.43\% & 69.03\% & 67.31\% & 63.16\% & 69.16\% & 62.78\% & 64.30\% & 69.10\% & 65.05\% \\
Self-Reflection & 69.79\% & \textbf{71.16\%} & 73.08\% & 70.29\% & 68.42\% & 67.43\% & 70.04\% & 69.79\% & 70.26\% \\
SE-Agent        & 64.14\% & 70.58\% & 67.25\% & 72.93\% & 68.81\% & 70.61\% & 68.54\% & 69.70\% & 68.93\% \\
AlphaEvolve     & 69.45\% & 69.50\% & 70.57\% & 71.07\% & 68.45\% & 67.66\% & 70.26\% & 68.98\% & 69.12\% \\
CSE (Ours)      & \textbf{71.17\%} & 70.39\% & \textbf{73.94\%}
                & \textbf{75.29\%} & \textbf{69.82\%} & \textbf{74.88\%} & \textbf{73.23\%} & \textbf{70.11\%} & \textbf{74.41\%} \\
\midrule

\multicolumn{10}{c}{\textbf{GPT-5}} \\
\midrule
Direct          & 60.21\% & 62.31\% & 61.66\% & 60.02\% & 62.60\% & 56.60\% & 60.12\% & 62.46\% & 59.13\% \\
Self-Reflection & 62.21\% & 63.09\% & 63.72\% & 61.79\% & 62.08\% & 56.76\% & 62.00\% & 62.59\% & 60.24\% \\
SE-Agent        & 61.74\% & 64.51\% & 65.36\% & 68.61\% & 62.70\% & 63.46\% & 65.18\% & 63.61\% & 64.41\% \\
AlphaEvolve     & 63.30\% & 63.96\% & 65.53\% & 64.13\% & 63.05\% & 60.54\% & 63.72\% & 63.51\% & 63.04\% \\
CSE (Ours)      & \textbf{65.46\%} & \textbf{66.82\%} & \textbf{68.10\%}
                & \textbf{69.94\%} & \textbf{64.42\%} & \textbf{66.83\%} & \textbf{67.70\%} & \textbf{65.62\%} & \textbf{67.47\%} \\
\bottomrule
\end{tabular}

\caption{\textbf{Main results on EffiBench-X (Python and C++).} ET, MP, and MI measure execution time, peak memory, and memory integral ratio relative to human solutions (higher is better). Avg is the per-metric mean across Python and C++. Best results are in bold.}
\label{tab:main_results}
\end{table*}

\subsection{Experimental Setup}
\label{sec:exp_setup}

\paragraph{Benchmark.}
We evaluate on EffiBench-X \cite{effibench}, which aggregates 623 algorithmic problems from major competitive programming platforms, including AtCoder, Codeforces, and LeetCode, spanning six programming languages. Each problem comes with strict time/memory limits and comprehensive test cases. We run experiments on two different languages (Python and C++) to assess the generality of our method. Following prior work \cite{effibench}, we report three normalized efficiency metrics: Execution-Time ratio (ET), Memory-Peak ratio (MP) and Memory-Integral ratio (MI), which compare the LLM-generated solution with the human reference on every problem.
Detailed definitions and derivations are provided in Appendix~\ref{app:metrics}.

\paragraph{Baselines.}
We compare CSE against three code-generation paradigms:
(1) Direct: a single-turn generation.
(2) Self-Reflection: iterative modification of the current best solution based on its feedback (e.g., variants of EffiLearner \cite{effilearner} and AfterBurner \cite{afterburner}).
(3) SE-Agent \cite{Se-agent}: a trajectory-level self-evolving agent method
(4) AlphaEvolve \cite{novikov2025alphaevolve}: a representative self-evolving agent method; we reproduce it using the open-source implementation of OpenEvolve \cite{openevolve}. 
We evaluate two open-source models (DeepSeek-V3-0324 \cite{deepseekv3} and Qwen3-235B-A22B \cite{qwen3}) and two closed-source models (Claude-4.5-Sonnet and GPT-5).

\paragraph{Implementation Details.}
All methods (except Direct) share a 30-candidate budget per task; the best solution encountered in the trajectory is recorded. Following EffiLearner's setup~\cite{effilearner}, efficiency is measured only on tasks already solved by Direct to decouple speed from correctness; if a method exhausts its budget without a valid solution, we fall back to the Direct baseline for that task, guaranteeing identical problem sets across evaluators. Across all methods, the reward is the raw memory--time integral, inverted into a maximization score to capture joint runtime and memory gains. See Appendix~\ref{app:impl} for more details.




\begin{figure}[t]
\centering

\begin{minipage}[t]{0.49\linewidth}
  \centering
  \includegraphics[width=\linewidth]{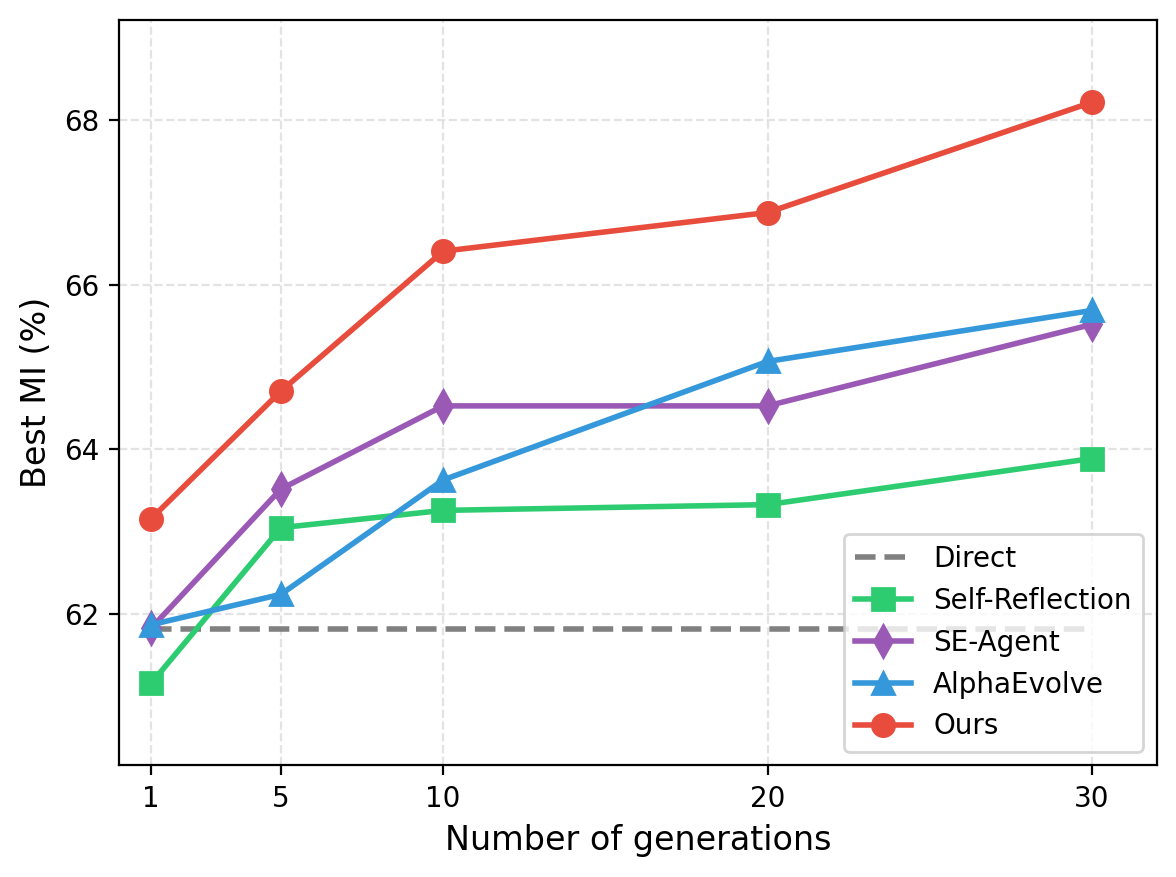}
  \captionof{figure}{\textbf{Best-so-far MI vs. generations.}
  At each generation $t$, we report the \emph{best-so-far} MI.}
  \label{fig:budget}
\end{minipage}\hfill
\begin{minipage}[t]{0.49\linewidth}
  \centering
  \includegraphics[width=\linewidth]{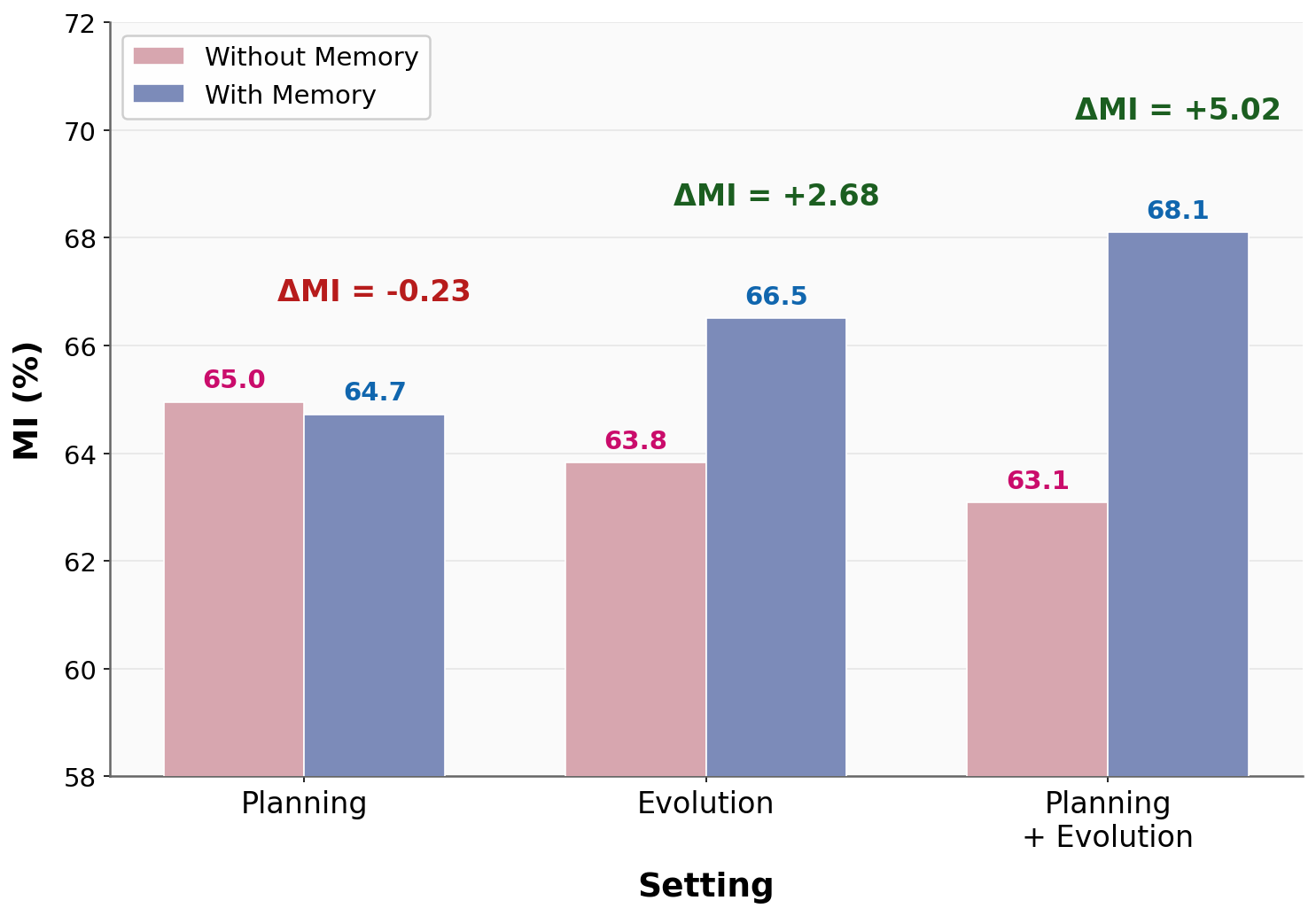}
  \captionof{figure}{\textbf{Synergy between \textit{Memory} and other modules.}
  We report MI and the gain $\Delta$MI from adding \textit{Memory} to each module combination.}
  \label{fig:synergy_memory_mi}
\end{minipage}

\end{figure}

\subsection{Main Results}

Table \ref{tab:main_results} presents the main experimental results on EffiBench-X across Python and C++ under a fixed test-time budget of 30 candidates per task. CSE consistently achieves the best performance across most three efficiency metrics (ET, MP, MI) on both programming languages, with particularly notable improvements on memory integral, demonstrating its effectiveness in discovering algorithmically efficient solutions within limited exploration budgets. The advantages hold across diverse model capabilities, from open-source models (Qwen3-235B-A22B, DeepSeek-v3-0324) to state-of-the-art closed-source models (Claude-4.5-Sonnet, GPT-5), validating that CSE's framework is model-agnostic and captures fundamental principles of efficient code evolution. Compared to population-based baselines AlphaEvolve and SE-Agent, CSE demonstrates clear superiority across most settings. This can be attributed to our key design that enables controlled evolution rather than uncontrolled stochastic exploration: through controlled mutation and compositional crossover, CSE achieves fine-grained, feedback-guided evolution that systematically navigate toward optimal solutions with higher exploration efficiency, while hierarchical memory allows the framework to leverage accumulated experiences from both previous iterations and related tasks to accelerate convergence and avoid repeated failures.

Beyond the final metrics, Figure~\ref{fig:budget} visualizes how performance evolves with increasing iteration budget.
CSE not only reaches a higher best-so-far MI, but also improves more rapidly in early generations and continues to make progress in later generations, suggesting stronger budget utilization under the same 30-generation constraint.

\subsection{Further Analysis}



\begin{table}[t]
\centering

\begin{minipage}[t]{0.49\linewidth}
  \centering
  \small
  \begin{tabular}{l r r r}
    \toprule
    \textbf{Setting} & \textbf{ET} & \textbf{MP} & \textbf{MI} \\
    \midrule
    \textbf{CSE (Full)} & \textbf{65.46\%} & \textbf{66.82\%} & \textbf{68.10\%} \\
    w/o \textit{Planning}        & 61.39\% & 64.40\% & 63.82\% \\
    w/o \textit{Evolution}       & 62.16\% & 64.28\% & 64.72\% \\
    w/o \textit{Memory}          & 59.87\% & 64.75\% & 63.08\% \\
    \bottomrule
  \end{tabular}
  \captionof{table}{\textbf{Ablation study.}
  \textit{Planning} means diversified planning initialization; \textit{Evolution} means genetic evolution; \textit{Memory} means hierarchical evolution memory.}
  \label{tab:ablation_python}
\end{minipage}\hfill
\begin{minipage}[t]{0.49\linewidth}
  \centering
  \small
  \begin{tabular}{lccc}
    \toprule
    \textbf{Method} & \textbf{\#Imp.} & \textbf{Iter@Best} & \textbf{Last-10 \#Imp.} \\
    \midrule
    SE-Agent     & 1.60 & 9.51  & 0.19 \\
    AlphaEvolve  & 0.90 & 7.44  & 0.06 \\
    \textbf{CSE (Ours)}   & \textbf{1.79} & \textbf{12.06} & \textbf{0.29} \\
    \bottomrule
  \end{tabular}
  \captionof{table}{\textbf{Statistics of evolution dynamics.}
  \textbf{CSE} achieves more frequent and more sustained improvements, indicating better-controlled exploration throughout evolution.
  }
  \label{tab:evo_dynamics}
\end{minipage}

\end{table}

\paragraph{Ablation Studies on CSE. } To investigate the individual contribution of each component in CSE, we conduct an ablation study where we systematically remove one component at a time while maintaining the same 30-candidate budget. Table~\ref{tab:ablation_python} presents the results. We evaluate three variants: CSE without \textit{Planning}, which eliminates the diversified initialization strategy; CSE without \textit{Evolution}, which removes the genetic operators; and CSE without \textit{Memory}, which discards the hierarchical evolution memory mechanism.
All three components prove essential, with \textit{Memory} showing the largest impact, followed by \textit{Planning} and \textit{Evolution}. The consistent performance drops across both ET and MP demonstrate that CSE's effectiveness arises from the synergistic interplay of all three components rather than any single mechanism.
We additionally provide qualitative case studies in Appendix~\ref{sec:ablation_case} to visualize the evolution dynamics under each ablation.

\paragraph{Analysis of \textit{Memory}.} 
To determine whether \textit{Memory}'s contribution is additive or conditional on other components, we compare different module combinations under the same 30-iteration budget. Fig ~\ref{fig:synergy_memory_mi} reports the MI and the gain from adding \textit{Memory} ($\Delta$MI).
The results reveal that \textit{Memory}'s effectiveness depends strongly on context. Adding \textit{Memory} to \textit{Planning} alone provides negligible benefit ($\Delta$MI = $-0.23$), while combining it with \textit{Evolution} yields substantial gains ($\Delta$MI = $+2.68$). The strongest effect emerges when all three components are present ($\Delta$MI = $+5.02$).
This demonstrates that \textit{Memory} amplifies controlled evolutionary processes rather than providing universal improvement. It is most valuable when the system can generate diverse candidates, refine them through evolution, and strategically revisit solutions, reinforcing guided exploration rather than offering standalone benefits.

\begin{figure*}[]
\centering
\includegraphics[width=\linewidth]{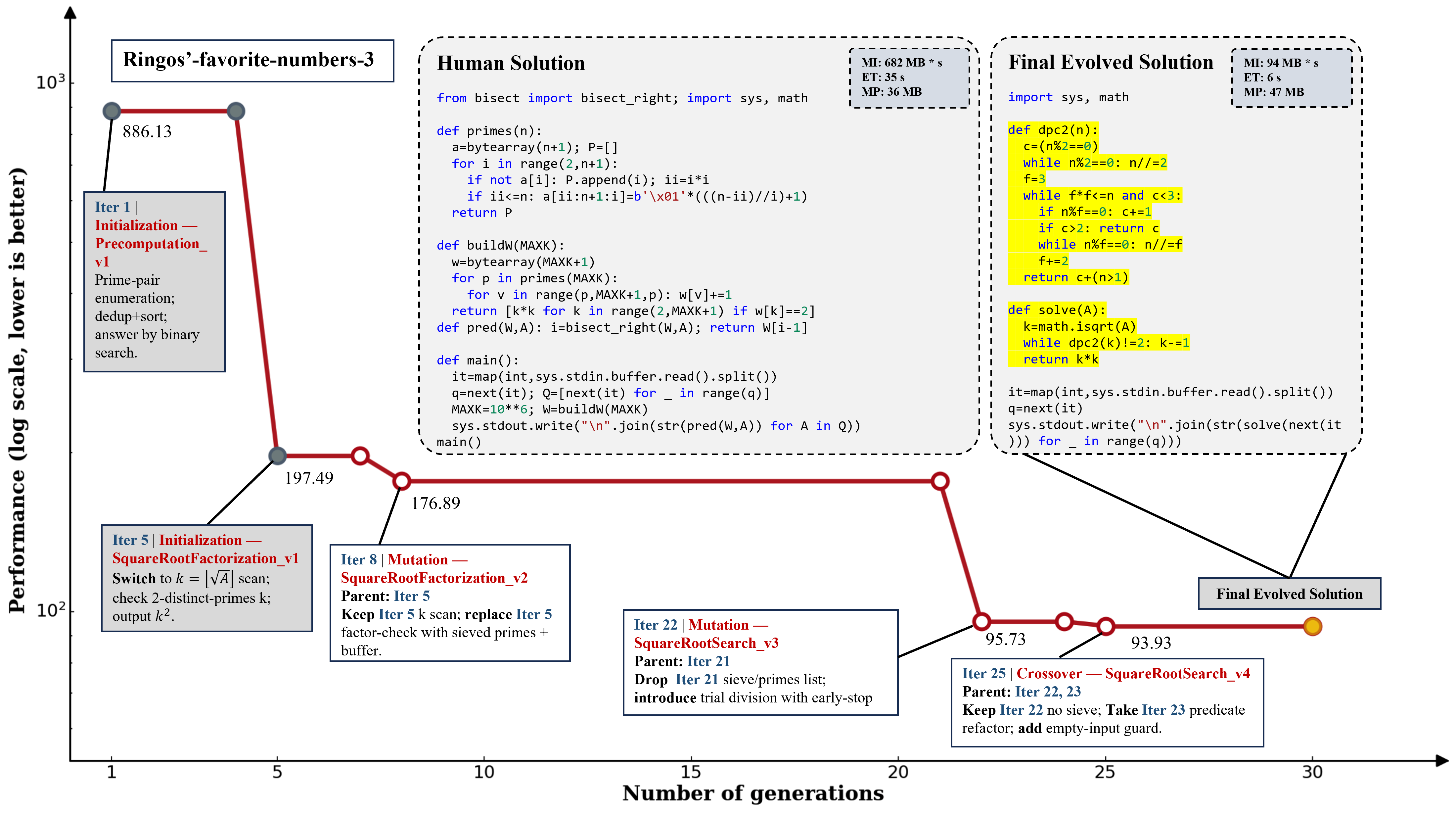}

\caption{\textbf{A Case Study of CSE evolution dynamics.}
To quantify progress, we plot the best-so-far \textbf{raw memory-time integral} (lower is better) against the number of generations.
The case highlights the concrete logic of key initialization, mutation, and crossover steps, and contrasts the final evolved solution against the human solution.}

\label{fig:case_overview}
\end{figure*}

\paragraph{Exploration on Evolution Dynamics.} 
To understand when and how frequently improvements occur during evolution, we conduct a fine-grained analysis over 30-iteration runs. Table~\ref{tab:evo_dynamics} reports three statistics: (i) the number of improvement times (\#Imp.), measuring how many iterations yield efficiency gains over the previous best; (ii) the iteration at which the final solution is found (Iter@Best), indicating whether the best result appears early or late; and (iii) improvements in the last 10 generations (Last-10 \#Imp.), quantifying the ability to sustain late-stage progress.
The results show that CSE achieves substantially more frequent improvements than baselines (\#Imp.\ = 1.79 vs.\ 0.90 for AlphaEvolve and 1.60 for SE-Agent) and notably stronger late-stage progress (Last-10 \#Imp.\ = 0.29 vs.\ 0.06 for AlphaEvolve and 0.19 for SE-Agent). This indicates that CSE continues exploring throughout the entire budget rather than plateauing early, contributing to its superior performance in Table~\ref{tab:main_results}.

\subsection{Case Study}

To demonstrate CSE in practice, Figure~\ref{fig:case_overview} shows the complete 30-iteration evolution on EffiBench-X \texttt{ringo's-favorite-numbers-3}. The trajectory illustrates how our mechanisms work together: initialization explores diverse solution hypotheses, including a precomputation-centric strategy (Iter 1) versus a reformulated
k-space search (Iter 5), showing CSE can switch high-level formulations rather than merely refining a single approach. Once the stronger structure emerges, mutations fix bottlenecks (e.g., inefficient checking or missing guards) while preserving the effective skeleton, yielding steady improvements. Crossover then merges complementary strengths from different parents, such as combining robust core search with cleaner predicates. Throughout, hierarchical evolution memory guides the search away from explored variants toward novel directions, reducing redundancy. The final solution differs markedly from the human reference in both design and implementation, achieving better efficiency.

\section{Conclusion}
We present Controlled Self-Evolution (CSE), a novel framework that addresses the low exploration efficiency of existing self-evolution methods for code optimization. By identifying three fundamental limitations, we propose CSE, including Diversified Planning Initialization, Genetic Evolution and Hierarchical Evolution Memory. Experiments on EffiBench-X demonstrate that CSE consistently outperforms all baselines across various LLM backbones, achieving higher efficiency from early generations while maintaining continuous improvement throughout evolution. Our work highlights the importance of controlled, feedback-driven exploration in self-evolution paradigms for code generation.

\section*{Limitations}
CSE can continuously improve solution quality through multi-round evolution.
However, we have not yet explored how to amortize this iterative optimization into the base model.
A promising yet underexplored direction is to distill CSE’s evolution trajectories into RL-style training \cite{DAPO,GSPO,codeRL} signals to strengthen the base model, enabling comparable or better optimization and producing higher-quality solutions.

\bibliography{custom}

\clearpage
\appendix

\section{Algorithmic Overview of CSE}

We summarize the end-to-end workflow of Controlled Self-Evolution (CSE) in Algorithm~\ref{alg:cse}.
CSE proceeds in three stages: (i) diversified planning initialization constructs a diverse initial population $\mathcal{P}_0$ by instantiating strategy sketches $\mathcal{Z}$; 
(ii) an iterative evolutionary loop performs reward-based parent selection, retrieves inter-task experiences from $\mathcal{M}^{\text{glb}}$ and injects them together with the intra-task memory $\mathcal{M}^{\text{loc}}_{t-1}$ into the context, then applies controlled mutation (\textsc{Refine}) or compositional crossover (\textsc{Crossover}) to generate $y^{(t)}_{\mathrm{child}}$;
(iii) after $T$ iterations, we return the best solution $y^*$ and distill task-level experience $g_\tau$ from top-$K$ improving/degrading steps to update $\mathcal{M}^{\text{glb}}$.
\begin{algorithm*}[t]
\caption{Controlled Self-Evolution (CSE)}
\label{alg:cse}
\begin{algorithmic}[1]
\REQUIRE Problem specification $x$, Global memory $\mathcal{M}^{\text{glb}}$, Max iterations $T$, retrieval queries $N_q$, top-$K$ size $K$
\ENSURE Optimized solution $y^*$

\STATE $\mathcal{Z} \leftarrow \text{PlanStrategies}(x)$
\STATE $\mathcal{P}_0 \leftarrow \{\, y_i^{(0)} \sim \mathcal{A}_{\theta}(y \mid x, z_i)\;|\; z_i \in \mathcal{Z}\,\}$
\STATE $\mathcal{M}^{\text{loc}}_0 \leftarrow \emptyset$
\STATE $\mathcal{S}_{\tau} \leftarrow \emptyset$

\FOR{$t = 1$ \TO $T$}
    \STATE Compute $\{\mathcal{F}(y,x)\}_{y \in \mathcal{P}_{t-1}}$
    \STATE Sample $y^{(t)}_{\mathrm{parent}} \leftarrow \text{SelectParent}(\mathcal{P}_{t-1})$ with $P_{\text{select}}(y)=\frac{\mathcal{F}(y,x)}{\sum_{y'\in\mathcal{P}_{t-1}}\mathcal{F}(y',x)}$
    \STATE $\{q_t^{(n)}\}_{n=1}^{N_q} \leftarrow \text{GenerateQueries}(\text{Context}_t)$
    \STATE $\mathcal{E}_t^{\text{ret}} \leftarrow \text{Retrieve}(\mathcal{M}^{\text{glb}}, \{q_t^{(n)}\}_{n=1}^{N_q})$
    \STATE $\text{Context}'_t \leftarrow \text{Compose}(\text{Context}_t, \mathcal{M}^{\text{loc}}_{t-1}, \mathcal{E}_t^{\text{ret}})$
    \STATE $\text{Op}_t \leftarrow \text{ChooseOp}(\text{Refine}, \text{Crossover})$
    \IF{$\text{Op}_t = \text{Refine}$}
        \STATE $y^{(t)}_{\mathrm{child}} \leftarrow \text{Refine}(y^{(t)}_{\mathrm{parent}} \mid \text{Context}'_t)$
    \ELSE
        \STATE Sample $y'^{(t)}_{\mathrm{parent}} \leftarrow \text{SelectParent}(\mathcal{P}_{t-1})$ with $P_{\text{select}}(\cdot)$
        \STATE $y^{(t)}_{\mathrm{child}} \leftarrow \text{Crossover}(y^{(t)}_{\mathrm{parent}}, y'^{(t)}_{\mathrm{parent}} \mid \text{Context}'_t)$
    \ENDIF
    \STATE $\Delta_t \leftarrow \mathcal{F}(y^{(t)}_{\mathrm{child}},x) - \mathcal{F}(y^{(t)}_{\mathrm{parent}},x)$
    \STATE $\mathcal{P}_t \leftarrow \text{UpdatePopulation}(\mathcal{P}_{t-1}, y^{(t)}_{\mathrm{child}}, \mathcal{F}(y^{(t)}_{\mathrm{child}},x))$
    \STATE $m_t \leftarrow \text{Reflect}(y^{(t)}_{\mathrm{child}}, y^{(t)}_{\mathrm{parent}}, \Delta_t)$
    \STATE $\mathcal{M}^{\text{loc}}_t \leftarrow \text{Compress}\!\left(\mathcal{M}^{\text{loc}}_{t-1}\cup\{m_t\}\right)$
    \STATE $\mathcal{S}_{\tau} \leftarrow \mathcal{S}_{\tau} \cup \{(y^{(t)}_{\mathrm{parent}}, y^{(t)}_{\mathrm{child}}, \Delta_t)\}$
\ENDFOR

\STATE $y^* \leftarrow \arg\max_{y \in \bigcup_{t=0}^{T} \mathcal{P}_t} \mathcal{F}(y, x)$
\STATE $\mathcal{S}^{+}_{\tau} \leftarrow \text{TopK}_{\max \Delta}(\mathcal{S}_{\tau}, K)$
\STATE $\mathcal{S}^{-}_{\tau} \leftarrow \text{TopK}_{\min \Delta}(\mathcal{S}_{\tau}, K)$
\STATE $g_{\tau} \leftarrow \text{Reflect}(\mathcal{S}^{+}_{\tau}, \mathcal{S}^{-}_{\tau}, y^*)$
\STATE $\mathcal{M}^{\text{glb}} \leftarrow \mathcal{M}^{\text{glb}} \cup \{g_{\tau}\}$

\RETURN $y^*$
\end{algorithmic}
\end{algorithm*}
\section{Evaluation Metrics.}
\label{app:metrics}
To quantify the efficiency of LLM-generated solutions relative to human-expert-written reference solutions, we follow prior work and report three normalized metrics: Execution Time (ET), Memory Peak (MP), and Memory Integral (MI).
We evaluate on $N$ problems, indexed by $i \in \{1,\dots,N\}$. For each problem $i$, we measure the execution time, peak memory usage, and memory--time integral for both the human reference solution (superscript $H$) and the LLM-generated solution (superscript $L$), under the same evaluation protocol.

\textbf{Failure Handling.}
If the LLM-generated solution for problem $i$ fails to pass all test cases or encounters a runtime error (e.g., timeout, crash), we assign a score of $0$ for that problem for all metrics.

\textbf{Clipping.}
To both (i) preserve the ability to credit solutions that outperform the human reference (ratios $>1$) and (ii) limit the influence of extreme outliers, we apply
\[
\mathrm{clip}(z,0,k)=\min(\max(z,0),k).
\]
In all experiments we set $k=5$ (shared across all methods), which retains sufficient headroom for surpassing human references while preventing a small number of extreme cases from dominating the average.

\textbf{(1) Execution Time (ET).}
Let $T_i^H$ denote the raw execution time of the human reference solution required to pass all test cases for problem $i$, and $T_i^L$ denote the execution time of the LLM-generated solution (conditioned on passing).
The per-problem ET score is
\begin{equation}
s_i^{T}=
\begin{cases}
0, & \text{if solution fails,}\\
\mathrm{clip}\!\left(\frac{T_i^{H}}{T_i^{L}},0,5\right), & \text{otherwise,}
\end{cases}
\end{equation}
and the overall ET is the mean score across problems, reported in a scaled form for readability:
\begin{equation}
\mathrm{ET}(\%)=\left(\frac{1}{N}\sum_{i=1}^{N} s_i^{T}\right)\times 100\%.
\end{equation}

\textbf{(2) Memory Peak (MP).}
Let $M_i^H$ and $M_i^L$ denote the raw peak memory usage of the human reference and the LLM-generated solution on problem $i$, respectively.
The per-problem MP score is
\begin{equation}
s_i^{M}=
\begin{cases}
0, & \text{if solution fails,}\\
\mathrm{clip}\!\left(\frac{M_i^{H}}{M_i^{L}},0,5\right), & \text{otherwise,}
\end{cases}
\end{equation}
and the overall MP is
\begin{equation}
\mathrm{MP}(\%)=\left(\frac{1}{N}\sum_{i=1}^{N} s_i^{M}\right)\times 100\%.
\end{equation}

\textbf{(3) Memory Integral (MI).}
To measure lifetime memory consumption, we define the raw memory--time integral for one execution as
\begin{equation}
A=\int_{0}^{T_{\text{total}}} m(t)\,dt,
\end{equation}
where $m(t)$ is the memory usage trace at time $t$ and $T_{\text{total}}$ is the total execution time. This integral is numerically approximated from high-resolution profiling traces.
Let $A_i^H$ and $A_i^L$ denote the memory integrals of the human reference and the LLM-generated solution on problem $i$, respectively.
The per-problem MI score is
\begin{equation}
s_i^{A}=
\begin{cases}
0, & \text{if solution fails,}\\
\mathrm{clip}\!\left(\frac{A_i^{H}}{A_i^{L}},0,5\right), & \text{otherwise,}
\end{cases}
\end{equation}
and the overall MI is
\begin{equation}
\mathrm{MI}(\%)=\left(\frac{1}{N}\sum_{i=1}^{N} s_i^{A}\right)\times 100\%.
\end{equation}

\section{Implementation Details}
\label{app:impl}

\paragraph{Experimental Setup Rationale.}
We follow the evaluation protocol used in EffiLearner~\cite{effilearner} to decouple \emph{efficiency} from \emph{correctness} when comparing iterative optimization methods.
In particular, efficiency is measured only on tasks that are already solved by \textsc{Direct}. This restriction ensures that all evaluated solutions are functionally correct, and avoids an apples-to-oranges comparison where different methods solve different subsets of tasks, making efficiency scores not directly comparable.
Moreover, if an iterative method exhausts the shared 30-candidate budget without producing a valid solution, we fall back to the \textsc{Direct} solution for that task, guaranteeing an identical set of evaluated problems across methods under a fixed budget.

\paragraph{Evaluation Protocol.}
Following our main experimental setup, each problem is evaluated on \textbf{100 test cases} with a time limit of \textbf{10\,s} and a memory limit of \textbf{1024\,MB}.
If a candidate program triggers a runtime error, exceeds the time limit (TLE), or violates the memory limit (MLE), we mark it as a \emph{failure} and assign it a reported metric of $0$ (for ET/MP/MI).

\paragraph{Raw Signal for Evolution.}
While the paper reports the \emph{normalized} metrics (ET/MP/MI) w.r.t.\ the human reference solution, the \emph{evolution process} is guided by the \textbf{raw memory--time integral} value, denoted as $A$ (lower is better).
To reduce measurement noise, we evaluate each candidate on the same machine for \textbf{5 repeated runs}; we remove outliers by discarding the maximum and minimum runs and take the mean of the remaining runs as the final $A$.
If any run results in failure (runtime error, TLE, or MLE), the candidate is treated as failed.

\paragraph{Reward and Parent Selection.}
Since smaller $A$ indicates better efficiency, we convert $A$ into a maximization-form reward for selection:
\[
\mathcal{F} \;=\; \frac{1}{A+\epsilon},
\]
where we set $\epsilon = 0.001$.
We sample parents proportional to their reward, i.e.,
\[
p_i = \frac{\mathcal{F}_i}{\sum_j \mathcal{F}_j}.
\]
Failed candidates are assigned $\mathcal{F}=0$, and thus will not be selected as parents.

\paragraph{Search Budget and Operator Schedule.}
We run evolution for $T=30$ iterations. At each iteration, we generate exactly \textbf{one} child candidate, resulting in a total budget of 30 evolved candidates per task.
We adopt a \textbf{deterministic alternating schedule} for the two evolution operators: \textbf{controlled mutation} and \textbf{compositional crossover} are executed in an interleaving order across iterations.
Concretely, we apply mutation on odd iterations and crossover on even iterations, ensuring a stable and reproducible operator mixture throughout evolution.

\paragraph{Initialization and Retrieval Hyperparameters.}
We use $N_{\text{init}}=5$ diverse initialization plans per problem.
For global memory retrieval, we generate $N_q=3$ queries and retrieve top-$K_m=3$ memories per query, resulting in at most $N_q \cdot K_m$ retrieved entries.

\paragraph{Embedding Model and Similarity Search.}
We encode both queries and memory entries using \textbf{Qwen3-8B-Embedding}.
We perform nearest-neighbor retrieval via \textbf{cosine similarity} and return the top-$K_m$ entries for each query.

\paragraph{Module Decomposition Template.}
To enable controlled mutation and crossover, we enforce a \textbf{fixed decomposition template} (in Figure \ref{fig:prompt_decomposition}) and require the LLM to output code in structured modules (e.g., \texttt{[A] Input}, \texttt{[B] Core}, \texttt{[C] Optimization module}).
The number of modules is fixed and determined by the template.
All subsequent mutation/crossover operations are applied at the module level, which helps maintain interface consistency and reduces unintended edits outside the targeted components.

\paragraph{Local Memory Compression.}
We maintain a local memory buffer for intra-task experiences.
To prevent memory overflow in the prompt context, we perform semantic compression whenever the local memory exceeds \textbf{1000 tokens}.

\paragraph{Global Memory Distillation and Construction Order.}
For each task $\tau$, we keep the top-$K$ improving and top-$K$ degrading steps (ranked by $\Delta$) to distill a task-level experience item $g_\tau$, with $K=5$.
To ensure determinism and reproducibility, we fix a single task-processing order and apply the same order to all methods when constructing and using the global memory.
After each processed task, we distill the evolved trajectory into compact experience items and add them into the global memory for retrieval in subsequent tasks.

\section{Ablation Case Studies}
\label{sec:ablation_case}

\begin{figure*}[t]
    \centering

    \begin{subfigure}{\linewidth}
        \centering
        \includegraphics[width=\linewidth]{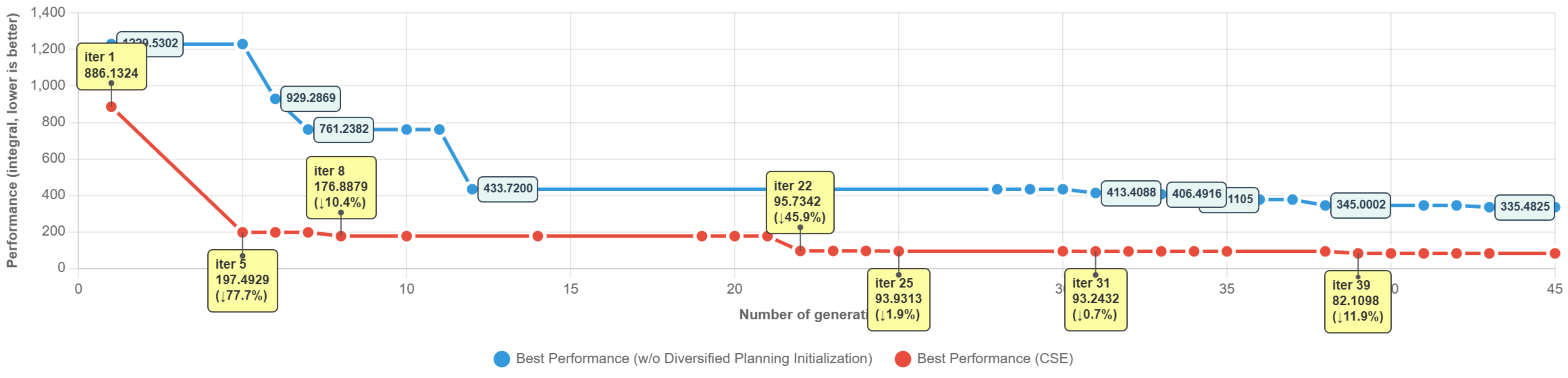}
        \caption{\textbf{w/o Diversified Planning Initialization.} Best-so-far performance over generations for full CSE (red) versus removing diversified planning initialization (blue).}
        \label{fig:ablation_init}
    \end{subfigure}

    \begin{subfigure}{\linewidth}
        \centering
        \includegraphics[width=\linewidth]{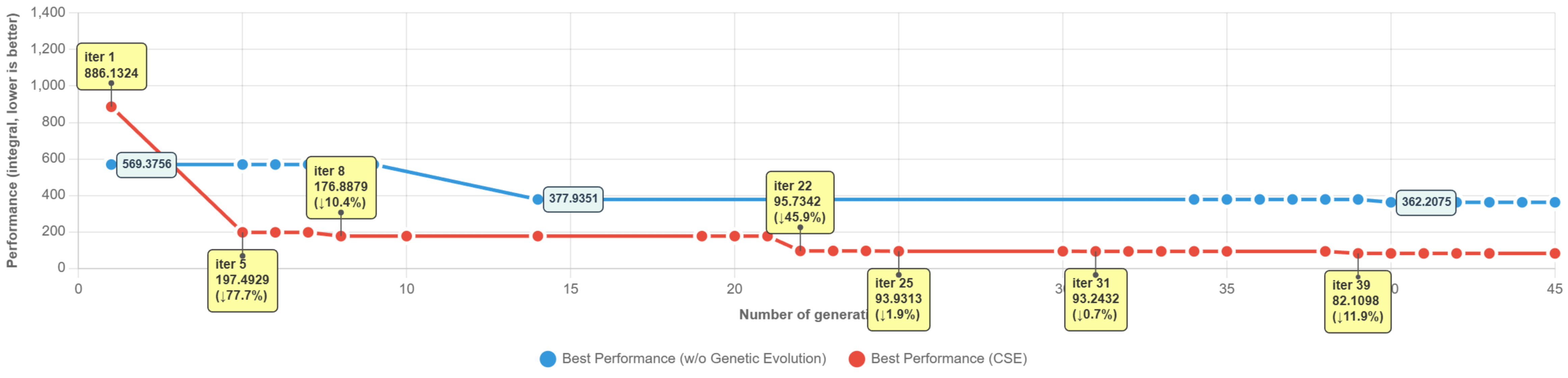}
        \caption{\textbf{w/o Genetic Evolution.} Best-so-far performance over generations for full CSE (red) versus removing genetic evolution (blue).}
        \label{fig:ablation_evo}
    \end{subfigure}

    \begin{subfigure}{\linewidth}
        \centering
        \includegraphics[width=\linewidth]{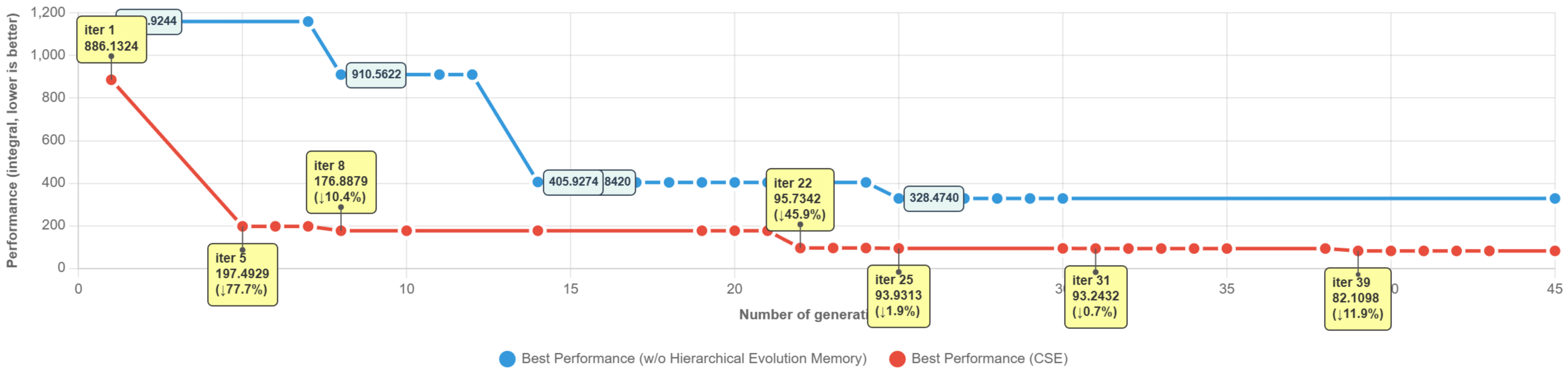}
        \caption{\textbf{w/o Hierarchical Evolution Memory.} Best-so-far performance over generations for full CSE (red) versus removing hierarchical experience memory (blue).}
        \label{fig:ablation_mem}
    \end{subfigure}

    \caption{\textbf{Case studies of ablation evolution dynamics.} We compare the best-so-far  \textbf{raw memory-time integral} (lower is better) across generations between full CSE (red) and three ablated variants (blue). Yellow callouts highlight representative milestone solutions along the CSE trajectory (e.g., \texttt{iter 1}, \texttt{iter 5}).}
    \label{fig:ablation_case}
\end{figure*}

\paragraph{Case Study: Diversified Planning Initialization.}
Figure~\ref{fig:ablation_init} compares full CSE with an ablated variant that removes diversified planning initialization. We observe a clear initialization bias: without diversified planning, the search starts from a substantially worse region (best-so-far $\approx$1230.5) and remains almost unchanged for several generations, indicating that the initial candidates are clustered around a low-quality mode. Although the ablated run eventually makes progress (dropping to $\approx$929.3 and $\approx$761.2 around mid-early generations), it still struggles to escape the basin and only reaches a plateau at $\approx$433.7 after $\sim$12 generations, with little improvement thereafter. In contrast, diversified planning immediately provides a much stronger and more promising starting point (iter~1: 886.1), enabling an early “algorithmic jump” within a few iterations (iter~5: 197.5, a 77.7\% reduction), followed by continued refinement (iter~8: 176.9). More importantly, starting from diverse, semantically distinct strategy sketches allows the evolutionary loop to explore multiple high-potential basins in parallel, which later supports a second breakthrough (iter~22: 95.7) and a final improvement (iter~25: 93.9). Overall, removing diversified planning causes prolonged stagnation and a much higher final plateau ($\approx$433.7 vs.\ 93.9), demonstrating that diversified planning is crucial for overcoming poor initial modes and accelerating the discovery of high-efficiency solution structures.

\paragraph{Case Study: Genetic Evolution.}
Figure~\ref{fig:ablation_evo} illustrates the effect of removing \emph{Genetic Evolution} (i.e., controlled mutation and compositional crossover). Although the ablated variant starts from a relatively strong initial point (best-so-far $\approx$569.4), it quickly becomes stagnant: the curve stays nearly flat for the first $\sim$10 generations and only achieves a modest improvement to $\approx$377.9 around the mid-run, after which it plateaus again until the end of the budget. In contrast, full CSE exhibits step-wise breakthroughs characteristic of effective evolutionary search: despite a worse starting best (iter~1: 886.1), it rapidly triggers a major early jump to iter~5 (197.5; $\downarrow$77.7\%), further refines to iter~8 (176.9), and later achieves another large improvement at iter~22 (95.7; $\downarrow$45.9\%) followed by a final gain at iter~25 (93.9). This gap suggests that the core benefit of Genetic Evolution is not merely producing more variants, but enabling \emph{structure-preserving, feedback-aligned} updates: controlled mutation can surgically repair the bottleneck component without disrupting already-correct parts (Eq.~\ref{eq:refine}), while compositional crossover can recombine complementary strengths across candidates rather than performing superficial text-level mixing (Eq.~\ref{eq:crossover}). Without these operators, the search lacks an effective mechanism to translate evaluation feedback into high-reward structural edits, leading to long plateaus and a much higher final performance floor ($\approx$377.9 vs.\ 93.9).

\paragraph{Case Study: Hierarchical Evolution Memory.}
Figure~\ref{fig:ablation_mem} examines the impact of removing \emph{Hierarchical Evolution Memory}. Without memory, the search trajectory exhibits pronounced plateaus and delayed progress: the best-so-far performance remains extremely high for multiple generations (around $\sim$1200), then only drops to $\sim$910.6, and later makes a larger decrease to roughly $\sim$405.9 before entering a long stagnation period near $\sim$400. Even late in the run, the memory-ablated variant only achieves a modest improvement to $\sim$328.5 and then plateaus again, indicating that exploration repeatedly revisits similar low-yield regions rather than systematically leveraging past lessons. In contrast, full CSE quickly consolidates early improvements (iter~5: 197.5; iter~8: 176.9) and, crucially, avoids long mid-run stagnation by triggering a late-stage breakthrough (iter~22: 95.7; $\downarrow$45.9\%) followed by a final refinement (iter~25: 93.9). This divergence highlights the role of memory as a feedback-to-action bridge: local memory distills parent--child reward changes into actionable success patterns and failure constraints (Eq.~\ref{eq:local}), reducing redundant trial-and-error within the task, while global memory retrieval supplies reusable optimization heuristics from prior tasks that steer generation toward unexplored but promising directions (Eq.~\ref{eq:global}). Removing this mechanism substantially slows the transition from incremental tweaks to high-impact structural changes, yielding a much higher final performance floor ($\sim$328.5 vs.\ 93.9).
\section{Prompt Templates}
\label{app:prompts}

This appendix section provides the exact system/user prompts used by our pipeline.
The templates cover diversified planning to propose multiple high-performance strategies; implementation prompts that synthesize a complete program under explicit constraints; slot-based decomposition prompts that diagnose correctness and efficiency by comparing a target solution against a reference baseline; evolution prompts for controlled mutation (reflection and refine) and compositional crossover (hybrid synthesis); and memory prompts that extract, compress, retrieve, and aggregate reusable success/failure experiences.
Across stages, prompts emphasize correctness-first optimization, favor algorithmic improvements over micro-optimizations, and optimize the raw memory--time integral to encourage balanced runtime--memory trade-offs.


\begin{promptbox}{System Prompt}
You are a world-class Algorithm Engineer. Design EXACTLY K distinct, high-performance strategies for the given problem.

Guidelines:
1. **Diversity**: Strategies MUST differ in algorithmic paradigms (DP, Greedy, Two Pointers, Bit Manipulation, etc.) or core data structures.
2. **Performance**: Prioritize optimal Time/Space complexity. Avoid brute-force unless necessary.
3. **Content**: Each strategy must include:
   - Core algorithmic logic
   - Key data structures
   - Expected Time/Space complexity (Big-O)

Output Format:
```json
{"strategies": ["<strategy_1>", "<strategy_2>", ...]}
```

The array must contain exactly K strings. NO extra text.
\end{promptbox}

\begin{promptbox}{User Prompt}
Problem Description:
{problem_text}

Required Strategy Count: {k}
\end{promptbox}

\begin{figure}[!htp]
\centering
\caption{Prompts for Diversified Planning Initialization.}
\label{fig:prompt_div_planning_init}
\end{figure}


\begin{promptbox}{System Prompt}
You are an expert competitive programmer specializing in algorithm optimization.

## Principles

1. **Correctness First**: Ensure correctness before optimizing for {optimization_target}.
2. **Algorithmic Focus**: Prefer algorithmic improvements over micro-optimizations. Explore directions **different from existing approaches**.
3. **Learn from Failures**: Use Memory to prune failed directions, but don't restrict creative exploration.
4. **Strict Improvement**: Your solution must outperform the best prior solution in `Additional Requirements`.

## Task

Synthesize a **complete program in {language}** that:

* Fixes fundamental bottlenecks in current solutions
* Explores novel optimization directions
* Minimizes the **integral** (area under memory-time curve) across all test cases — balancing both runtime and memory

## Problem

{task_description}

## Response Format

### 1. Thinking

* **Baseline Analysis**: Identify strategy mode and best baseline from `Additional Requirements`.
* **Failure Pruning**: Review failed directions from Memory; avoid repeating them.
* **New Direction**: Select a fundamentally different, high-potential approach. Justify why it's better for {optimization_target}.
* **Risks**: Note correctness concerns and trade-offs.

### 2. Final Code

Complete, self-contained {language} program. No test harness or debug prints.

## Allowed Imports

{allowed_imports_scope}

## Context

**Additional Requirements**:

### STRATEGY MODE: PLAN STRATEGY

You must strictly follow and implement the outlined approach below.

{strategy}

**Local Memory**: {local_memory}
**Global Memory**: {global_memory}
\end{promptbox}

\begin{promptbox}{User Prompt}
Using all context (Additional Requirements, Local/Global Memory, current program), generate `## Thinking` and `## Final Code` to improve PERFORMANCE. Follow the guidance from Memory.
\end{promptbox}

\begin{figure}[!htp]
\centering
\caption{Prompts for Implement Sketches.}
\label{fig:prompt_implement_sketches}
\end{figure}


\begin{promptbox}{System Prompt}
You are a code diagnosis agent. Analyze **Target Solution** for a {language} problem, decompose into slots, and diagnose correctness/performance vs **Reference Solution**.

**Optimization target**: {optimization_target} (integral = area under memory-time curve). Balance runtime and memory.

### Task

1. **Summarize**: Nickname + high-level strategy
2. **Factorize**: 3-5 slots (io_parsing, core_logic, edge_case, perf_patch, misc)
3. **Diagnose**: Identify bottlenecks, risks, good parts to inherit

### Output JSON

```json
{
"solution_name": "String",
"approach_summary": "String",
"slot_view": {
  "slots": [{"slot_id": "io_parsing|core_logic|edge_case|perf_patch|misc", "description": "String", "tags": ["impl_method"], "code_span": [start, end]}],
  "diagnoses": [{"slot_id": "String", "status": "ok|bottleneck|bug_source|risky|redundant", "correctness_level": "good|minor_risk|major_risk|unknown", "perf_level": "strong|weak|neutral|unknown", "priority": "inherit|optimize|inspect|low", "evidence": ["..."]}]
}
}
```

### Rules

* **tags**: Use specific impl methods (e.g., "segment_tree", "fast_io"), not generic descriptions
* **perf_level**: Compare complexity vs Reference (O(N²) vs O(N) → weak)
* **priority**: inherit=keep, optimize=bottleneck, inspect=risky
* **Self-contained evidence**: Be specific (e.g., "O(N²) nested loop" not "slower than Reference")

Return JSON only (can wrap in ```json fence).
\end{promptbox}

\begin{promptbox}{User Prompt}
Diagnose the Target Solution vs Reference.

## Problem

{problem_description}

## Reference Solution (Baseline)

{best_solution}

## Target Solution

{target_solution}

Generate JSON: decompose into slots, assign tags, set perf_level vs Reference, set priority (optimize bottlenecks, inherit good parts).
\end{promptbox}

\begin{figure}[!htp]
\centering
\caption{Prompts for Decomposition (slot-based diagnosis).}
\label{fig:prompt_decomposition}
\end{figure}


\begin{promptbox}{System Prompt}
You are an expert competitive programmer specializing in algorithm optimization.

## Principles

1. **Correctness First**: Ensure correctness before optimizing for {optimization_target}.
2. **Algorithmic Focus**: Prefer algorithmic improvements over micro-optimizations. Explore directions **different from existing approaches**.
3. **Learn from Failures**: Use Memory to prune failed directions, but don't restrict creative exploration.
4. **Strict Improvement**: Your solution must outperform the best prior solution in `Additional Requirements`.

## Task

Synthesize a **complete program in {language}** that:

* Fixes fundamental bottlenecks in current solutions
* Explores novel optimization directions
* Minimizes the **integral** (area under memory-time curve) across all test cases — balancing both runtime and memory

## Problem

{task_description}

## Response Format

### 1. Thinking

* **Baseline Analysis**: Identify strategy mode and best baseline from `Additional Requirements`.
* **Failure Pruning**: Review failed directions from Memory; avoid repeating them.
* **New Direction**: Select a fundamentally different, high-potential approach. Justify why it's better for {optimization_target}.
* **Risks**: Note correctness concerns and trade-offs.

### 2. Final Code

Complete, self-contained {language} program. No test harness or debug prints.

## Allowed Imports

{allowed_imports_scope}

## Context

**Additional Requirements**:

### STRATEGY MODE: REFLECTION AND REFINE

You must reflect on the previous solution and implement targeted improvements.

### SOURCE SOLUTION SUMMARY

{source_summary}

### REFINEMENT GUIDELINES

1. **Diagnose**: Identify main shortcomings (correctness risks, performance bottlenecks, redundant work, I/O overhead).
2. **Targeted Fixes**: Apply code-level changes to the faulty component ONLY:

   * Algorithmic upgrade for the specific bottleneck
   * Data structure replacement for the problematic module
   * Caching/precomputation for repeated work
3. **Preserve Working Parts**: Keep the remaining well-performing components frozen.
4. **Correctness First**: Ensure correctness before optimizing runtime.

**Local Memory**: {local_memory}
**Global Memory**: {global_memory}
\end{promptbox}

\begin{promptbox}{User Prompt}
Using all context (Additional Requirements, Local/Global Memory, current program), generate `## Thinking` and `## Final Code` to improve PERFORMANCE. Follow the guidance from Memory.
\end{promptbox}

\begin{figure}[!htp]
\centering
\caption{Prompts for Controlled Mutation.}
\label{fig:prompt_controlled_mutation}
\end{figure}


\begin{promptbox}{System Prompt}
You are an expert competitive programmer specializing in algorithm optimization.

## Principles

1. **Correctness First**: Ensure correctness before optimizing for {optimization_target}.
2. **Algorithmic Focus**: Prefer algorithmic improvements over micro-optimizations. Explore directions **different from existing approaches**.
3. **Learn from Failures**: Use Memory to prune failed directions, but don't restrict creative exploration.
4. **Strict Improvement**: Your solution must outperform the best prior solution in `Additional Requirements`.

## Task

Synthesize a **complete program in {language}** that:

* Fixes fundamental bottlenecks in current solutions
* Explores novel optimization directions
* Minimizes the **integral** (area under memory-time curve) across all test cases — balancing both runtime and memory

## Problem

{task_description}

## Response Format

### 1. Thinking

* **Baseline Analysis**: Identify strategy mode and best baseline from `Additional Requirements`.
* **Failure Pruning**: Review failed directions from Memory; avoid repeating them.
* **New Direction**: Select a fundamentally different, high-potential approach. Justify why it's better for {optimization_target}.
* **Risks**: Note correctness concerns and trade-offs.

### 2. Final Code

Complete, self-contained {language} program. No test harness or debug prints.

## Allowed Imports

{allowed_imports_scope}

## Context

**Additional Requirements**:

### STRATEGY MODE: CROSSOVER

Synthesize a SUPERIOR hybrid solution by combining the best elements of two prior trajectories.

### SOLUTION 1 SUMMARY

{solution_1}

### SOLUTION 2 SUMMARY

{solution_2}

### SYNTHESIS GUIDELINES

1. **Complementary Combination**: Actively combine specific strengths from each solution.

   * If T1 has better core algorithm but slow I/O, and T2 has fast I/O but weaker algorithm → implement T1's algorithm with T2's I/O.
   * If T1 has correct logic but slow structure, and T2 has fast structure but buggy logic → implement T1's logic with T2's structure.
2. **Avoid Shared Weaknesses**: If both failed at a specific sub-task, introduce a novel fix.
3. **Seamless Integration**: The result must be a single, cohesive implementation—not concatenated code.

**Local Memory**: {local_memory}
**Global Memory**: {global_memory}
\end{promptbox}

\begin{promptbox}{User Prompt}
Using all context (Additional Requirements, Local/Global Memory, current program), generate `## Thinking` and `## Final Code` to improve PERFORMANCE. Follow the guidance from Memory.
\end{promptbox}

\begin{figure}[!htp]
\centering
\caption{Prompts for Compositional Crossover.}
\label{fig:prompt_compositional_crossover}
\end{figure}


\begin{promptbox}{System Prompt}
You observed an evolutionary step where metrics **improved**. Extract strategy-level insights.

Goal:

1. Decide if this is truly **Success** or just **Neutral** (noise/trivial refactor).
2. If strategy-level changes exist, extract up to 3:

   * **Direction items**: Reusable optimization strategies
   * **Memory items**: Reasoning patterns explaining WHY it works

Strategy-level changes include: algorithm switch, data structure change, I/O optimization, major loop restructuring.
Do NOT create items for: variable renaming, formatting, measurement noise.

Output Format:

```json
{
  "thought_process": "Brief reasoning.",
  "new_direction_items": [
    {"direction": "Strategy name", "description": "When/why to use.", "status": "Success | Neutral"}
  ],
  "new_memory_items": [
    {"type": "Success", "title": "Technique name", "description": "One-sentence summary", "content": "2-6 sentences on when/why it works."}
  ]
}
```

\end{promptbox}

\begin{promptbox}{User Prompt}

## Source Solutions

{source_solutions}

## Current Solution

{current_solution}

## Current Directions (Strategy Board)

{directions}
\end{promptbox}

\begin{figure}[!htp]
\centering
\caption{Prompts for Local Memory Extract (Success).}
\label{fig:prompt_local_memory_success}
\end{figure}


\begin{promptbox}{System Prompt}
You observed an evolutionary step where metrics **regressed** or correctness broke. Extract warnings.

Goal:

1. Decide if this is truly **Failure** or just **Neutral** (noise).
2. If strategy-level changes caused the regression, extract up to 3:

   * **Direction items**: Strategies to mark as Failed/risky
   * **Memory items**: Anti-patterns explaining WHY it failed

For Failure memory items:

* Title: Describe the SPECIFIC mistake (e.g., "BFS without boundary check causes OOB"), not just "BFS implementation"
* Content: Explain what went wrong and what condition triggered it

Output Format:

```json
{
  "thought_process": "Brief reasoning.",
  "new_direction_items": [
    {"direction": "Failed strategy", "description": "Why problematic.", "status": "Failed | Neutral"}
  ],
  "new_memory_items": [
    {"type": "Failure", "title": "Avoid ...", "description": "Why dangerous", "content": "What went wrong, how to avoid."}
  ]
}
```

\end{promptbox}

\begin{promptbox}{User Prompt}

## Source Solutions

{source_solutions}

## Current Solution (Failed)

{current_solution}

## Current Directions (Strategy Board)

{directions}
\end{promptbox}

\begin{figure}[!htp]
\centering
\caption{Prompts for Local Memory Extract (Failure).}
\label{fig:prompt_local_memory_failure}
\end{figure}


\begin{promptbox}{System Prompt}
Compress the **direction_board** of an evolutionary agent.

Rules:

1. **Merge similar strategies**: Combine entries describing the same idea.
2. **Do NOT merge different failure modes**: Keep separate if root causes differ.
3. **Aggregate counts**: When merging, SUM success_count and failure_count; update status accordingly.
4. **Prune low-value entries**: Remove vague or noise entries. Keep ~5-10 useful directions.

Output Format:

```json
{
  "thought_process": "Brief explanation.",
  "direction_board": [
    {"direction": "Strategy", "description": "Explanation", "status": "Success|Failed|Neutral", "success_count": N, "failure_count": M}
  ]
}
```

Compress the **experience_library** of an evolutionary agent.

Rules:

1. **Merge overlapping experiences**: Combine entries describing the same lesson.
2. **Do NOT merge different root causes**: "Avoid recursion without memoization (TLE)" and "Avoid large array (OOM)" are DIFFERENT.
3. **Content Guidelines**:

   * Success: Explain WHY it works, under what conditions
   * Failure: Describe the SPECIFIC mistake and what triggered it
4. **Filter trivial items**: Remove noise entries. Keep ~5-8 useful experiences.

Output Format:

```json
{
  "thought_process": "Brief explanation.",
  "experience_library": [
    {"type": "Success|Failure", "title": "Specific title", "description": "One-sentence summary", "content": "2-6 sentences on when/why."}
  ]
}
```

\end{promptbox}

\begin{promptbox}{User Prompt}

## Directions (Strategy Board)

{directions}

## Experiences

{experience_library}
\end{promptbox}

\begin{figure}[!htp]
\centering
\caption{Prompts for Local Memory Compress.}
\label{fig:prompt_local_memory_compress}
\end{figure}


\begin{promptbox}{System Prompt}
Design search queries for a global memory of past optimization experiences.

Your job:

1. Read the context (problem, constraints, language, optimization target, local memory).
2. Think about potential bottlenecks in this task.
3. Output 2-3 concrete queries to search for useful experiences.

Query format: Short natural-language descriptions of scenario + problem, e.g.:

* "In Python, how to handle fast IO when reading 2e5 integers"
* "For pair counting with N up to 2e5, how to avoid O(N^2) loops"
* "How to reduce DP memory when table is O(N^2) and N=5000"

Output Format:

```json
{
  "thought_process": "Brief analysis of key bottlenecks.",
  "queries": ["<query_1>", "<query_2>", "<optional_query_3>"]
}
```

\end{promptbox}

\begin{promptbox}{User Prompt}

## Problem Description

{problem_description}

## Optimization Target

{optimization_target}

## Language

{language}

## Local Memory

{local_memory}
\end{promptbox}

\begin{figure}[!htp]
\centering
\caption{Prompts for Query Generation (memory retrieval).}
\label{fig:prompt_query_generation}
\end{figure}


\begin{promptbox}{System Prompt}
Extract generalizable experiences from optimization trajectories for the SAME problem.

You will receive:

* **Improvement Steps**: Changes that improved metrics
* **Regression Steps**: Changes that worsened metrics
* **Best Solution**: Current best solution

Goal: Build an experience library with Success/Failure lessons reusable on future tasks.

Guidelines:

1. **Contrastive reasoning**: Compare improvements vs regressions. What consistently works/fails?
2. **Strategy-level focus**: Algorithm choice, data structure, I/O patterns, caching. Ignore cosmetic changes.
3. **Generalize**: Abstract away specific names. Describe for problem families (e.g., "pair counting with constraints", "DP with large N").
4. **Limit**: Output at most 3-5 experiences. Prioritize high-impact, clearly explainable changes.

Output Format:

```json
{
  "thought_process": "Brief contrastive analysis.",
  "experiences": [
    {
      "type": "Success | Failure",
      "title": "For Success: technique name. For Failure: 'Avoid ...'",
      "description": "One-sentence summary.",
      "content": "1-5 sentences on when/why this works or fails."
    }
  ]
}
```

\end{promptbox}

\begin{promptbox}{User Prompt}

## Problem Description

{problem_description}

## Improvement Steps

{improvement_steps}

## Regression Steps

{regression_steps}

## Best Solution

{best_solution}
\end{promptbox}

\begin{figure}[!htp]
\centering
\caption{Prompts for Global Memory Extract.}
\label{fig:prompt_global_memory_extract}
\end{figure}

\end{document}